\pdfoutput=1
\documentclass[letterpaper, 10pt, conference]{ieeeconf}      % Use this line for a4 paper
\IEEEoverridecommandlockouts 
\usepackage{caption}

\usepackage{amsmath}
\usepackage{amssymb}
\usepackage{cite}
\usepackage[pdftex]{graphicx} 
\usepackage{algorithm} % Algorithm Library 
\usepackage{algorithmic} % Algorithm Library 
\usepackage{multirow} % Table 
\usepackage{tabularx} % Table \usepackage{hhline} % Table 
\usepackage{array} % Align Package 
\usepackage{url} % Hyperlink package. (e.g. \url{~~~~}) 
\usepackage{fixltx2e} % float package \usepackage{stfloats} % float package 
\usepackage{dblfloatfix}

\title{\LARGE \bf
Keeping Less is More: Point Sparsification for Visual SLAM
}
\author{Yeonsoo Park$^{1}$ and Soohyun Bae$^{2}$% <-this % stops a space
\thanks{$^{1}$Yeonsoo Park is with Mobiltech, Republic of Korea. Email: {\tt\footnotesize yspark@mobiltech.io}}%
\thanks{$^{2}$Soohyun Bae is with Bobidi, USA. Email: {\tt\footnotesize soohyun@bobidi.com}}%
}
\begin{document}

\maketitle
\thispagestyle{empty}
\pagestyle{empty}

%%%%%%%%%%%%%%%%%%%%%%%%%%%%%%%%%%%%%%%%%%%%%%%%%%%%%%%%%%%%%%%%%%%%%%%%%%%%%%%%
\begin{abstract}
When adapting Simultaneous Mapping and Localization (SLAM) to real-world applications, such as autonomous vehicles, drones, and augmented reality devices, its memory footprint and computing cost are the two main factors limiting the performance and the range of applications. In sparse feature based SLAM algorithms, one efficient way for this problem is to limit the map point size by selecting the points potentially useful for local and global bundle adjustment (BA). This study proposes an efficient graph optimization for sparsifying map points in such SLAM systems. 
% The proposed sparsification process simultaneously considers multiple aspects of the pose-graph optimization. 
Specifically, we formulate a maximum pose-visibility and maximum spatial diversity problem as a minimum-cost maximum-flow graph optimization problem. The proposed method works as an additional step in existing SLAM systems, so it can be used in both conventional or learning based SLAM systems. By extensive experimental evaluations we demonstrate the proposed method achieves even more accurate camera poses with approximately 1/3 of the map points and 1/2 of the computation.

\end{abstract}

%%%%%%%%%%%%%%%%%%%%%%%%%%%%%%%%%%%%%%%%%%%%%%%%%%%%%%%%%%%%%%%%%%%%%%%%%%%%%%%%
\section{INTRODUCTION}

Simultaneous Localization and Mapping (SLAM) has been extensively studied for a wide range of applications such as indoor mapping~\cite{Schueftan15}, drone~\cite{Stumberg17}, self-driving vehicles~\cite{qin20}, virtual reality, and augmented reality~\cite{Lepetit17}. Advances in computing systems and elaborate sensor technologies in cameras and LiDAR have accelerated the SLAM system adaptions in those applications. 
Especially, visual SLAM is one of the most frequently used system for mapping since it can be embedded on any device with a low-cost vision sensor.
%Most of the visual SLAM is build on graph optimization concept and can be divided into two groups: 
Most of the visual SLAM systems are based on graph optimization concept and can be divided into two groups: 
sparse feature-based visual SLAM extracts feature points from an image and tracks them in image sequences to calculate their camera poses and generate a three-dimensional map. Then, the positions of the landmarks constituting the 3D map are re-projected with the estimated poses of cameras and updated to minimize the distance from the coordinates of the feature points tracked from the image~\cite{klein07PTAM, murartal17orbslam2}. 
%Direct SLAM method is to minimize the difference of intensity from the next image acquired when the first image is converted to an image at the second location in order to obtain three-dimensional information on camera movement and environment from two images~\cite{engel14,fontan20}. 
Direct SLAM is to minimize the difference of pixel intensity from the next image acquired when the first image is converted to an image at the second location in order to obtain three-dimensional information on camera movement and environment from two images~\cite{engel14,fontan20}. 
It has advantages in homogeneous environment where insufficient local features extracted, but in many cases feature-based SLAM is preferred for real-time performance due to its superior processing speed and computational efficiency.
\begin{figure}[ht!]
\centering
\includegraphics[width=8cm]{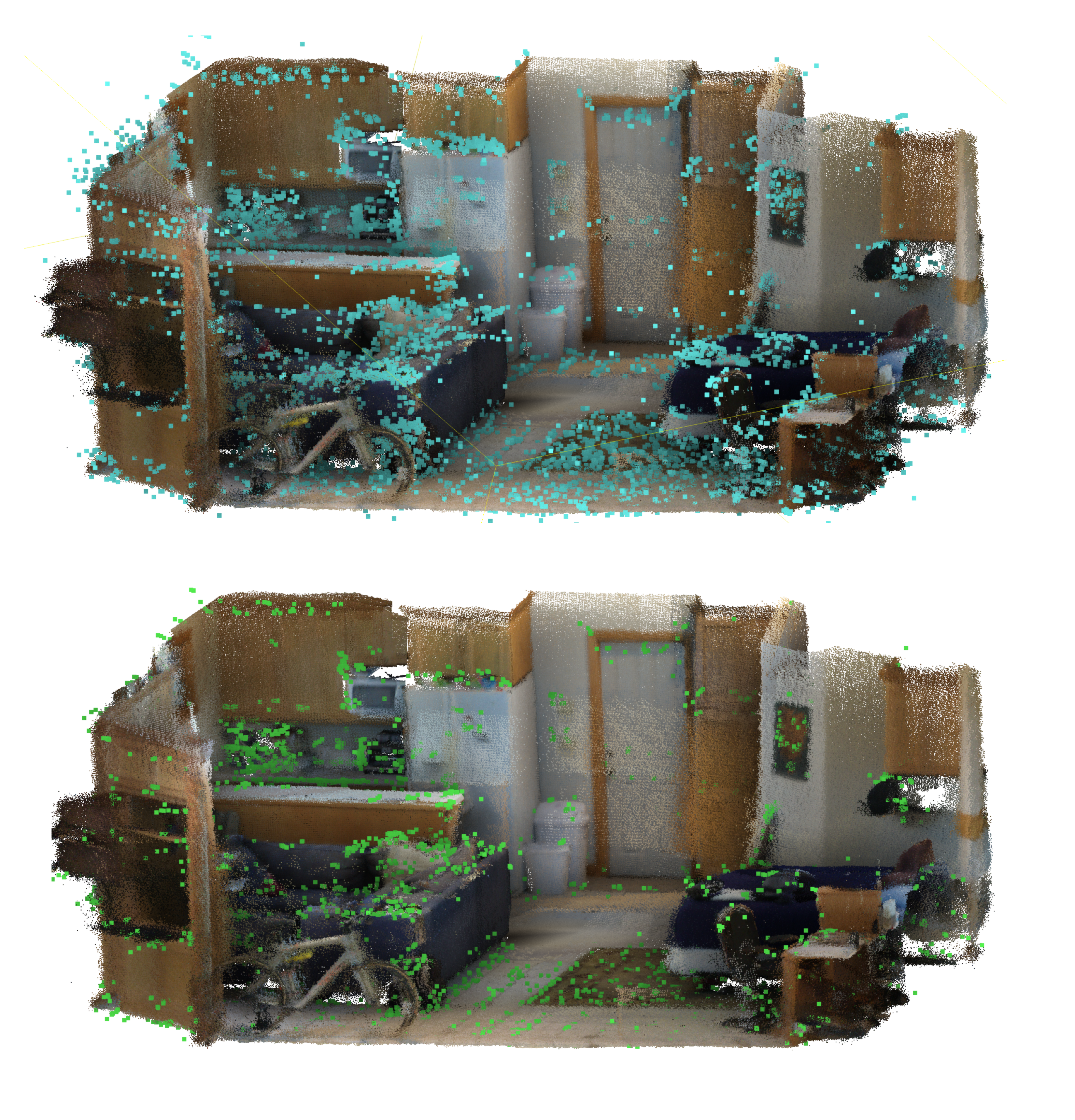}
      \caption{{\tt\bf Top:} map points generated by ORB-SLAM2. {\tt\bf Bottom:} sparsified map with proposed method. Estimated pose accuracy is marginally improved both on building map and performing localization on the map. Map at the bottom consists of 39\% of the points compared to the top.}
      \label{figure_door}
\end{figure}
However, one of the biggest pitfalls of the visual SLAM systems is its quadratically increasing memory size and computation cost as the map size grows. To tackle such growing resource requirements, a series of efforts have been made along two directions: one way is to solve the optimization problems efficiently~\cite{zhou20sba}. 
Many algorithms have tried to reduce the computation cost by utilizing the map topology or the problem structure. 
The other way is to reduce the problem size throughout the whole SLAM system including feature/frame selection, keyframe/3D point decimation, and so on. Most of them focus on reducing either the data size or computation cost while mildly sacrificing the pose accuracy. To reduce the map size and computation cost simultaneously in an existing SLAM system while maintaining the pose localization accuracy, we introduce an efficient point sparsification algorithm that can be incorporated directly into any feature-based visual SLAM pipeline. 
Our contributions include:
\begin{itemize}
\item Proposes a graph representation of the camera pose pairs and 3D points for maximum point visibility
\item Proposes a new cost for maximizing the spatial diversity of the 2D features on the image space
\item Proposes a minimum-cost maximum-flow based point sparsification algorithm for controlling the remaining number of points
\item Provides a detailed pose accuracy, point reduction, and speed improvement comparison with various indoor / outdoor public datasets.
\end{itemize}
%To best of our knowledge, this is the first work of integrating multiple properties regarding to the feature and relationship with frames at once to sparsify feature map, and also the first to provide the verification of maintenance in localization performance for the sparsified map.
To the best of our knowledge, this is the first work of integrating multiple properties regarding to the feature and relationship with frames at once to sparsify feature map, and also the first to provide the verification of maintenance in localization performance for the sparsified map.

\section{RELATED WORK}
As visual SLAM has been becoming an active area, there have been a series of research on fast computation of the optimization problems in SLAM and on problem space reduction for a low memory and computation demand. They can be roughly divided into two areas. 

First, graph based optimizations have been studied for a fast pose optimization~\cite{concha19}. Joan {\it et al.} ~\cite{vallve18,vallve2019pose} improves the accuracy and the speed of SLAM by reducing iterative optimization of KLD (Kullback-Leibler Divergence) using the factor descent and noncyclic factor descent. %Paull {\it et al.}~\cite{paull19} formulates the node selection problem over the penalty to be paid in the resulting sparsification. 
Paull {\it et al.}~\cite{paull19} formulates the node selection problem by minimizing the sparsification penalty.
Using the distribution of nodes, the KLD is minimized by selecting the optimal set of subnodes by approximating the dense distribution to the sparse distribution. Huang {\it et al.}~\cite{huang13} sparsifies nodes through a marginalization of old nodes while maintaining all information about the remaining nodes, and formulating a normalized minimization problem to keep the graph composition sparse. Wang {\it et al.}~\cite{wang18AprilSAM} devises a method for reordering dynamic variables and reducing the work associated with inverse permutations for the fast incremental Cholesky factorization to decide between incremental and batch updates, providing computational savings for incremental SLAM algorithms. %Frey~\cite{frey18} proposes a heuristic method that reduces computations for both batch and incremental optimization through an elimination complexity metric that bridges the analytical gap between the graph structures and its computations. 
Frey~\cite{frey18} proposes elimination complexity (EC) metric, an analysis tool that interprets the relationship between global graph structure and computation, and shows that simple decimation/keyframing through the proposed metric can achieve great computational efficiency.
Hsiung {\it et al.}~\cite{hsiung18} proposes the fixed-lag method that marginalizes variables in the SLAM problem and minimizes information loss during the graph sparsification. %Choudhary {\it et al.}~\cite{choudhary15} uses an information-based approach and the incremental version of the minimization problem to efficiently sparse the number of landmarks and poses without compromising the accuracy of the estimated trajectory.
Choudhary {\it et al.}~\cite{choudhary15} uses an information-based approach and the incremental version of the minimization problem to efficiently sparsify the number of landmarks and poses while maintaining the accuracy of the estimated trajectory.

The other group of methods reduces the graph geometry in SLAM, which decimates features~\cite{cvisic15}, points~\cite{concha19,dias19}, frames~\cite{lin19,fanfani16} with minimal information loss. Bailo {\it et al.}~\cite{bailo18anms} proposes an adaptive non-maximal suppression (ANMS) to quickly and uniformly re-segment keypoints in the image. 
%It reduces the computational complexity through a square approximation of the search range to suppress irrelevant points, and initializes the search range based on the image dimensions, which leads to a faster convergence. 
It reduces the computational complexity by suppressing irrelevant points through a square approximation of the search range, and leads to a faster convergence by initializing the search range according to the image dimension.
Gauglitz {\it et al.}~\cite{gauglitz11} efficiently selects a spatially distributed set of keypoints through the suppression via disk covering (SDC) algorithm that clusters keypoints based on an approximated nearest neighbor and the greedy approach. Opdenbosch {\it et al.}~\cite{Opdenbosch18} proposes a strategy to extract useful features by referencing multiple frames by utilizing the temporal correlation between successive frames and weighting features through the tracking in the SLAM system.

\section{PROPOSED METHOD}

We first review an existing visual SLAM system where our proposed method is integrated for evaluation. 
Once the connectivity among map points, estimated by triangulating rays from $n>1$ frames, and camera poses is represented as a graph structure with flow capacities and costs, we present a solution of the graph representation for point sparsification.

\subsection{ORB-SLAM2 Revisit}
Since ORB-SLAM2~\cite{murartal17orbslam2} has been proposed, it has been used as a reference visual SLAM method because of its real-time tracking performance and improved loop closure accuracy on mono, stereo cameras, and even RGB-D sensors. So we evaluate the performance of the proposed method on ORB-SLAM2, the proposed can be easily adapted in any feature based mono or stereo visual SLAM though.
\begin{figure*}[h!]
\includegraphics[width=18cm]{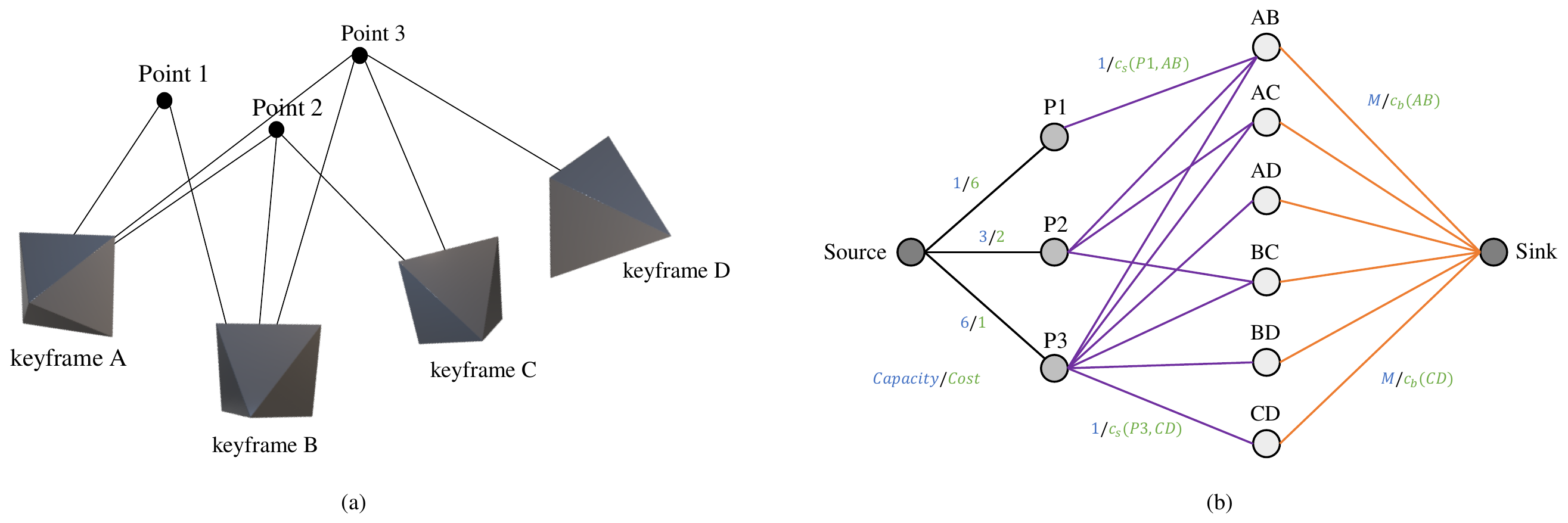}
  \caption{{\tt\bf(a)} An example of four keyframes (A, B, C, and D) that share three points in the 3D space.
{\tt\bf(b)} constructed bipartite graph from (a). There are 4 layers of vertices: two special vertices, a set of point vertices, and a set of frame pair vertices. At the above of each edge, Capacity/Cost value is written for the example case.}
  \label{figure_dia}
  \vspace{-0.3cm}
\end{figure*}

One of the key factors that affect the memory and computation requirements in ORB-SLAM2 and in the visual SLAM system is the number of map points and local interest point features associated with the map points. As they grow, the size of local and global BA increases quadratically, which in turn requires significantly increased computation cost. So, two main ways have been studied: 1) extracting only relevant local features \& points, or 2) decimating such features or points that does not contribute much to the pose optimization. In the proposed method, we focus on decimating points to build a more simplified BA problem after extracting enough interest points and generating map points temporarily and demonstrate that the proposed method significantly improve performances by extensive evaluations.

\subsection{Graph Representation for Point Sparsification}

Fig. \ref{figure_dia} shows an example of a simplified local map structure to be optimized in bundle adjustments. We build a directed flow graph structure to interpret the relationship among frames and points viewed by them. In this example, there are 4 keyframes connected together and 3 points they are observing. For efficiently selecting the points among the frame pairs, we configure this point-side and frame-side connectivity as a bipartite graph $G=(v, e)$ structure which includes source, sink, and two different set of vertices as Fig. \ref{figure_dia} (b). The core problem that the proposed method tackles is how to select a subset of the points that have minimal structural changes onto the local and global bundle adjustment problems. It is equivalent to how to select such points that maximizes the number of constraints in the BA problem while minimizing the number of points. 
In addition, for adjusting 6DOF pose for each frame, the residual errors on the image space forms the error covariance matrix of each camera poses, so the evenly distributed residual errors help make the covariances are well regularized. Simply speaking, if all the interest points are grouped near a corner of the image space, the pose adjustment problem space has steep edges, so it becomes harder to solve efficiently. Similarly, a wider frame baseline between two frames makes the Jacobian vales of the constraints on the points are similar. Intuitively, if the baseline angle of the two frames for one point is near zero, the depth estimation becomes harder and the point adjustment also becomes harder. The three observations can be summarized as the following three goals in the point selection problem:
\begin{itemize}
    \item Maximum point visibility: the number of frames shares one point is maximized.
    \item Maximum spatial diversity: the distribution of interest points on the image space is diversified.
    \item Maximum frame baseline length: the camera center distance between any two frames are maximized.
\end{itemize}

To solve the aforementioned problems in one integrated algorithm, we propose a new method based on a directed graph representation where nodes correspond to points and pose pairs. In this graph, the costs and flow capacities among the nodes are used for turning the actual point visibility, spatial diversity, and the baselines of the nodes into a minimum-cost maximum-flow bipartite graph. In this bipartite graph, a vertex on point-side, which corresponds to one of the $m$ points in the map, is notated as $v_{p_i}$ where $p_i\in P=\{p_1, p_2, ... , p_m\}$ and a set of point vertices is represented as $\mathbf{V}_P$. Vertices on the next layer represent any possible frame pairs connected via $\mathbf{V}_P$. A frame pair vertex $v_{f_{ij}} \in \mathbf{V}_F$ indicates the pair of $f_i$ and $f_j$ where $f_i, f_j\in F=\{f_1, f_2, ... f_k\}$, where $k$ is the number of keyframes in the map. We notate the source vertex which has only outgoing edges that are connected to $\mathbf{V}_P$ as $v_{so}$. For the sink vertex which has only incoming edges connected from $\mathbf{V}_F$, we use the notation $v_{si}$. Under this configuration, we assign the costs and capacities quantifying desirable attributes for each of the edges.

\subsubsection{Point Connectivity}

First, we consider the connectivity between one point and the frames that shares the interest points of the point. 
The point with a high connectivity indicates high visibility with robust local features. 
Such highly visible points are prone to be selected since they provide strong constraints across multiple poses on pose graph. So, we define the cost function $c_c$ for the edges between $v_{so}$ and points $v_{p_i}$ as below to be lower value of cost for $v_{p_i}$ with high connectivity:
%$$c(e(v_{so}, v_{p_i})) = cost_m(n) \eqno{(1)}$$
\begin{align}
    c_c(e(v_{so}, v_{p_i})) = & \notag \\
    c_m(n) = & \left\lceil (n+1)/(n - 1) \cdot c_m(n + 1)\right\rceil
    \label{eq:cost_point}
\end{align}
, where $n$ denotes the number of frames viewing the point $p_i$ associated with the vertex $v_{p_i}$ connected to the source edge and $m$ is the maximum number of $n$ on the $v_P$. 
The function $c_m(n)$ is determined according to the value of $m$ and computed recursively where $c_m(m)$ has the value of $1$.
For the edge $e_p(v_{so}, v_{p_i})$, the capacity is simply set to $n(n-1)/2$, the number of edges connected between $v_{p_i}$ and $v_F$ which is equal to the number of frame pairs viewing the point $p_i$.

\subsubsection{Spatial Diversity of Interest Points}

Many studies have reported that the use of spatially homogeneous set of features improves image registration performance ~\cite{gruber2010optimization, cheng2006determining, brown2005multi}. 
For the same goal ORB-SLAM2 includes a process of selecting interest points for spatially homogeneous distribution during the ORB feature extraction step. However, an initial homogeneous distribution of interest points does not guarantee a similar distribution of interest points in the following steps including point sparsification. Thus, we define the spatial cost $c_s$ for the edge $e(v_P, v_F)$ to ensure that the feature distribution is maintained or even improved during the point sparsification.
\begin{align}
  c_s(e(v_{p_i}, v_{f_{jk}}))= \left\lfloor \log_{10}(N(p_i, f_j) \cdot N(p_i, f_k)+1) \right\rfloor
  \label{eq:cost_spatial_div}
\end{align}
, where $N(p_i, f_j)$ denotes the number of nearby keypoints $p_i$ on the frame $f_j$. We consider keypoints are nearby if the keypoints exist in the box centered on the reference keypoints with a fixed size. We set the size of the searching box as $64 \times 48$, equivalent to the grid size that ORB-SLAM2 used for feature extraction. The capacity of these edges is set to one.

\subsubsection{Frame Pair Baseline}

Lastly, we consider the baseline distance of each frame pair. An optimization on a point observed in the frame pair that are in more than a certain distance can be done more reliably, and potentially useful for compensating the drift error accumulated among keyframes. With the purpose of preserving points which provide valuable constraint in pose graph optimization, baseline cost $c_b$ is applied to edges between $v_F$ and $v_{si}$:
\begin{align}
c_b(e(v_{f_{jk}}, v_{si})) = \left\lceil \cfrac{10}{0.1 \cdot d(f_j, f_k) + 1} \right\rceil
\label{eq:cost_frame_baseline}
\end{align}
, where $d(f_j, f_k)$ is the $L_2$ norm of $(O_{f_j}, O_{f_k})$ and $O_{f_j}$ is the camera center of the frame $f_j$. Here, the amount of flow that goes through $e(v_{f_{jk}}, v_{si})$ is equivalent to the number of points that are shared between $f_j$ and $f_k$. So, by constraining the maximum flow on this edge with the capacity $M$, we can control the desired number of points for each frame pair. The smaller $M$, the fewer points selected.

\subsection{Minimum Cost Maximum Flow Graph Optimization} %부제 수정 필요

We solve the aforementioned graph problem using a minimum cost maximum flow algorithm~\cite{edmonds72} to compute the maximum flow from $v_{so}$ to $v_{si}$ that minimizes the total cost:
\begin{align}
C = \sum_{e}{f(e)c(e)}
\label{eq:total_cost}
\end{align}
, where $f(e)$ is flow on the edge $e$. By computing the optimum flow with the minimum cost under the constraint of capacity, we measure the flow on the edges between the points and the frames with respect to the degree to which the three desired conditions defined above are satisfied. After computing flow, we only take the point $p_i$ whose flow on edge $e(v_{so}, v_{p_i})$ is larger then a pre-set threshold $\theta_{f}$. Goldberg’s algorithm ~\cite{goldberg1997efficient} guarantees that the worst case time complexity is bounded by $O(n^2m \cdot \log(n \cdot C))$, where $n$ is the total number of vertices and $m$ is the total number of edges, and $C$ is the biggest input cost. 

\section{EXPERIMENTAL RESULTS}

In this section, we evaluate and demonstrate the performance of the proposed method. We first present the implementation details of the proposed method. Then, we provide an analysis of experimental results performed extensively on various datasets, including observations of interrelated multiple factors we propose by comparing with other existing methods.

\begin{table}[]
\centering
%\caption{Selected point and keyframe ratio according to the parameter $M$ on EuRoC sequences}
\caption{Selected point and keyframe ratio on EuRoC sequences}
\label{table_euroc}
\resizebox{\columnwidth}{!}{%
\begin{tabular}{l|rr|rr|rr|rr}
\hline
 & \multicolumn{2}{l|}{Original} & \multicolumn{2}{l|}{Ours ($M=100$)} 
 & \multicolumn{2}{l|}{Ours ($M=200$)} & \multicolumn{2}{l}{Ours ($M=300$)} \\ \hline
Sequence & \multicolumn{1}{l}{\#MPs} & \multicolumn{1}{l|}{\#KFs} & \multicolumn{1}{l}{MP (\%)} & \multicolumn{1}{l|}{KF (\%)} & \multicolumn{1}{l}{MP (\%)} & \multicolumn{1}{l|}{KF (\%)} & \multicolumn{1}{l}{MP (\%)} & \multicolumn{1}{l}{KF (\%)} \\ \hline\hline
MH01 & \multicolumn{1}{r}{19,534} & 398 & 23.96 & 34.42 & \multicolumn{1}{r}{56.24} & 61.81 & \multicolumn{1}{r}{85.70} & 85.18 \\ \hline
MH02 & \multicolumn{1}{r}{17,397} & 342 & 25.48 & 38.30 & \multicolumn{1}{r}{57.28} & 65.20 & \multicolumn{1}{r}{82.10} & 82.16 \\ \hline
MH03 & \multicolumn{1}{r}{21,699} & 418 & 20.70 & 34.93 & \multicolumn{1}{r}{51.89} & 62.44 & \multicolumn{1}{r}{80.87} & 87.80 \\ \hline
MH04 & \multicolumn{1}{r}{18,461} & 297 & 26.05 & 60.61 & \multicolumn{1}{r}{52.76} & 79.12 & \multicolumn{1}{r}{77.85} & 86.53 \\ \hline
MH05 & \multicolumn{1}{r}{19,389} & 318 & 22.58 & 50.31 & \multicolumn{1}{r}{51.62} & 65.09 & \multicolumn{1}{r}{80.05} & 85.85 \\ \hline
V101 & \multicolumn{1}{r}{9,194} & 123 & 43.68 & 76.42 & \multicolumn{1}{r}{74.53} & 92.68 & \multicolumn{1}{r}{93.78} & 98.37 \\ \hline
V102 & \multicolumn{1}{r}{12,034} & 158 & 40.15 & 74.68 & \multicolumn{1}{r}{75.32} & 96.84 & \multicolumn{1}{r}{93.26} & 98.10 \\ \hline
V103 & \multicolumn{1}{r}{21,657} & 260 & 37.14 & 66.54 & \multicolumn{1}{r}{67.42} & 85.00 & \multicolumn{1}{r}{74.50} & 86.15 \\ \hline
\end{tabular}
}
\vspace{-0.3cm}
\end{table}

\begin{figure}[ht!]
\centering
\includegraphics[width=9cm]{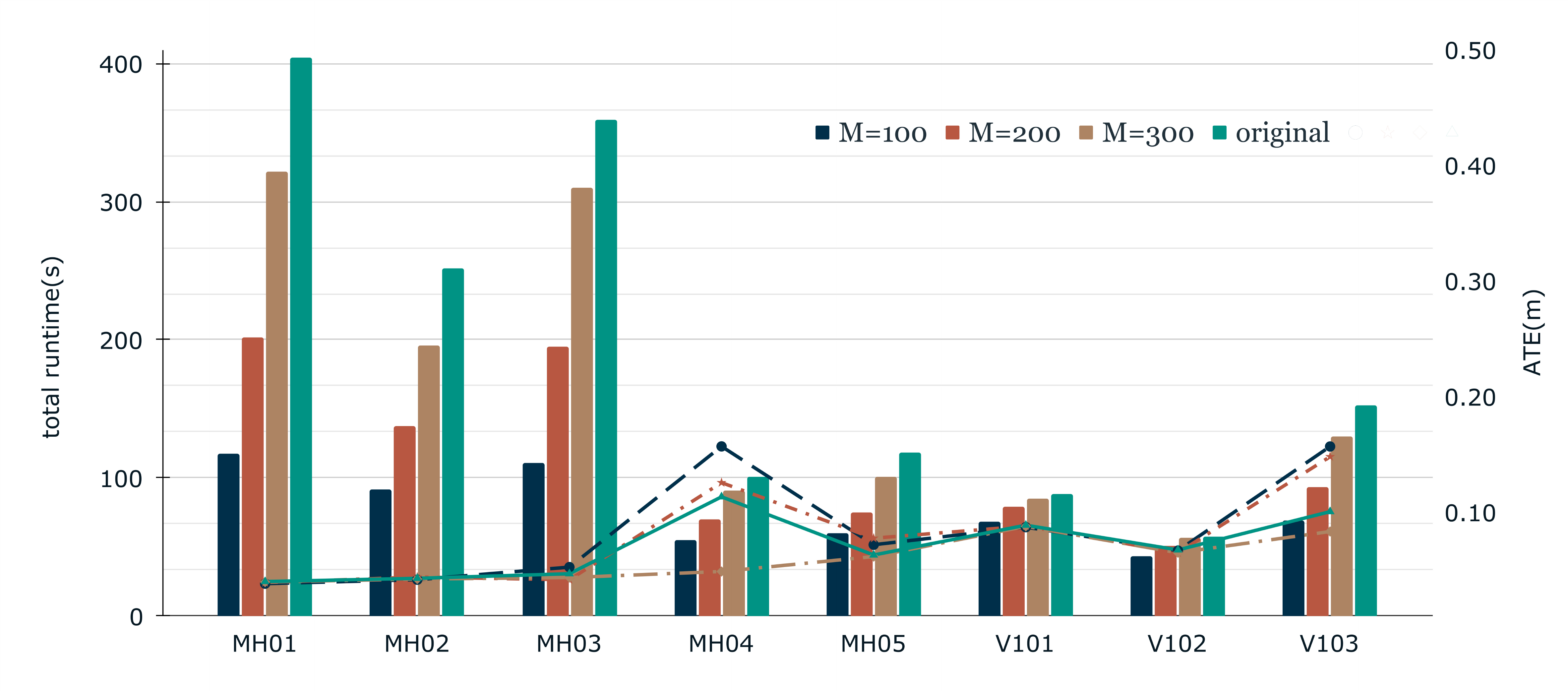}
  \caption{Total runtime and RMS ATE on EuRoC related to Table \ref{table_euroc}.}
  \label{figure_graph}
  \vspace{-0.3cm}
\end{figure}

\begin{table*}[t!]
\centering
\caption{Comparison of the proposed method across multiple $M$. Sequences are TUM, ICL-NUIM, and ScanNet from the top. RMS ATE are expressed as relative differences from the original result. Error values lower than original are in bold.}

\label{table_rgbd}
\resizebox{\textwidth}{!}{
\begin{tabular}{l|rrrrr|rrrrr|rrrrr} \hline 
& \multicolumn{5}{c|}{Ours ($M=100$)} & \multicolumn{5}{c|}{Ours ($M=140$)} & \multicolumn{5}{c}{Ours ($M=200$)} \\ \hline
Sequence 
& MP (\%) & KF (\%) & time (\%) & ATE$_r$ ($^{\circ}$) & ATE (m) 
& MP (\%) & KF (\%) & time (\%) & ATE$_r$ ($^{\circ}$) & ATE (m) 
& MP (\%) & KF (\%) & time (\%) & ATE$_r$ ($^{\circ}$) & ATE (m) \\ \hline\hline
fr1\_desk 
    & 41.89 & 71.88 & 69.02 & \textbf{-0.07640} & \textbf{-0.00098} 
    & 56.09 & 81.25 & 77.39 & \textbf{-0.16145} & 0.00000
    & 75.20 & 92.71 & 90.62 & 0.02042 & \textbf{-0.00083} \\
fr1\_desk2 
    & 48.28 & 82.61 & 78.17 & 0.49822 & \textbf{-0.00207} 
    & 56.75 & 85.22 & 81.22 & 0.47606 & 0.00230
    & 85.28 & 91.30 & 89.89 & \textbf{-0.35920} & \textbf{-0.00004} \\
fr2\_desk 
    & 35.78 & 63.28 & 75.22 & 0.12407 & \textbf{-0.00313} 
    & 53.74 & 79.66 & 82.19 & 0.17511 & \textbf{-0.00308} 
    & 78.00 & 92.66 & 91.71 & 0.01537 & 0.00043 \\
f2\_xyz 
    & 46.92 & 45.71 & 87.28 & 0.09626 & 0.00143 
    & 59.31 & 65.71 & 90.65 & 0.01762 & 0.00033 
    & 82.77 & 88.57 & 94.81 & \textbf{-0.00208} & 0.00034 \\
fr3\_office 
    & 23.54 & 45.85 & 58.92 & 0.02044 & 0.00020 
    & 36.08 & 57.64 & 62.55 & 0.03020 & \textbf{-0.00023} 
    & 55.13 & 73.36 & 71.60 & \textbf{-0.02061} & \textbf{-0.00422} \\
fr3\_nt 
    & 42.27 & 94.74 & 92.03 & \textbf{-0.15584} & \textbf{-0.01066} 
    & 50.91 & 97.37 & 90.43 & \textbf{-0.15153} & \textbf{-0.00150} 
    & 65.18 & 100.00 & 93.82 & \textbf{-0.17051} & \textbf{-0.01306} \\ \hline
lr\_kt0 
    & 34.59 & 55.56 & 65.75 & \textbf{-0.00890} & 0.00114
    & 47.25 & 61.11 & 69.22 & 0.06359 & 0.00125 
    & 64.10 & 76.67 & 76.47 & 0.04053 & 0.00043 \\
lr\_kt1 
    & 51.86 & 76.60 & 87.41 & \textbf{-1.90292} & \textbf{-0.02725} 
    & 63.35 & 82.98 & 89.11 & \textbf{-2.08678} & \textbf{-0.01543} 
    & 77.51 & 97.87 & 92.76 & \textbf{-1.99993} & \textbf{-0.01552} \\
lr\_kt2 
    & 34.45 & 47.06 & 73.37 & \textbf{-0.06925} & 0.00295 
    & 45.16 & 57.35 & 76.53 & \textbf{-0.52515} & 0.00231
    & 59.28 & 72.06 & 82.76 & \textbf{-0.49117} & \textbf{-0.00189} \\
lr\_kt3 
    & 44.90 & 78.95 & 94.27 & 0.08689 & 0.00097
    & 59.42 & 88.16 & 96.19 & \textbf{-0.17873} & \textbf{-0.00061} 
    & 77.09 & 93.42 & 100.94 & \textbf{-0.18718} & \textbf{-0.00034} \\
of\_kt0
    & 35.45 & 43.42 & 68.64 & 0.21269 & 0.00112
    & 46.12 & 52.63 & 74.00 & 0.29504 & 0.00093
    & 62.96 & 68.42 & 81.65 & 0.46068 & 0.00183 \\
of\_kt1 
    & 39.99 & 52.24 & 80.19 & 0.06706 & 0.00188 
    & 50.84 & 58.21 & 83.71 & \textbf{-0.00193} & \textbf{-0.01078} 
    & 72.21 & 53.73 & 99.28 & 0.00782 & \textbf{-0.00151} \\
of\_kt2 
    & 40.30 & 39.35 & 86.45 & 0.02681 & 0.00856 
    & 55.22 & 43.88 & 88.16 & 0.03672 & 0.00675
    & 49.25 & 65.27 & 95.31 & 0.01088 & 0.00458 \\
of\_kt3 
    & 60.23 & 94.44 & 89.80 & 0.58354 & \textbf{-0.00034} 
    & 55.51 & 100.00 & 92.35 & \textbf{-0.16572} & \textbf{-0.00351} 
    & 88.43 & 100.00 & 99.47 & \textbf{-0.20288} & \textbf{-0.00232} \\ \hline
0000\_00 
    & 38.55 & 59.05 & 68.95 & 0.20591 & \textbf{-0.00875} 
    & 56.42 & 75.86 & 71.17 & 0.09285 & 0.00259 
    & 75.90 & 89.01 & 86.66 & 0.16438 & \textbf{-0.00826} \\
0009\_00 
    & 10.60 & 11.84 & 8.18 & 0.37079 & 0.00020
    & 14.66 & 14.75 & 14.64 & 0.12933 & \textbf{-0.00218} 
    & 37.60 & 40.62 & 23.87 & \textbf{-0.02666} & 0.00135 \\ \hline\hline
Avg. 
    & 39.35 & 60.16 & 73.98 & 0.00496 & \textbf{-0.00217} 
    & 50.43 & 68.86 & 77.50 & \textbf{-0.12217} & \textbf{-0.00130} 
    & 69.12 & 80.98 & 85.73 & \textbf{-0.17126} & \textbf{-0.00244} \\ \hline
\end{tabular}
}
\end{table*}

\begin{table*}[]
\centering
\caption{Experimental result on outdoor environment using KITTI in stereo modes}
\label{table_kitti}
\resizebox{\textwidth}{!}{
\begin{tabular}{l|rrrrr|rrrrr|rrrrr}\hline
 & \multicolumn{5}{c|}{Original} & \multicolumn{5}{c|}{Ours ($M=100$)} & \multicolumn{5}{c}{Ours ($M=200$)} \\ \cline{2-16} 
Sequence 
& \multicolumn{1}{l}{\#MPs} & \multicolumn{1}{l}{\#KFs} & \multicolumn{1}{l}{time (s)} & \multicolumn{1}{l}{ATE$_r$ ($^{\circ}$)} & \multicolumn{1}{l|}{ATE (m)} 
& \multicolumn{1}{l}{MP (\%)} & \multicolumn{1}{l}{KF (\%)} & \multicolumn{1}{l}{time (\%)} & \multicolumn{1}{l}{ATE$_r$ ($^{\circ}$)} & \multicolumn{1}{l|}{ATE (m)} 
& \multicolumn{1}{l}{MP (\%)} & \multicolumn{1}{l}{KF (\%)} & \multicolumn{1}{l}{time (\%)} & \multicolumn{1}{l}{ATE$_r$ ($^{\circ}$)} & \multicolumn{1}{l}{ATE (m)} \\ \hline\hline
07 
    & 26,789 & 235 & 52.67 & 0.53871 & 0.5367
    & 36.2 & 98.7 & 87.03 & \textbf{0.51575} & \textbf{0.5076} 
    & 70.8 & 100.0 & 94.58 & \textbf{0.48100} & \textbf{0.4903} \\
08 
    & 95,649 & 1,011 & 197.56 & \textbf{1.25505} & 3.2719
    & 42.7 & 99.3 & 90.41 & 1.48866 & 3.5912 
    & 82.4 & 100.0 & 96.08 & 1.35641 & \textbf{3.2391} \\
10 
    & 41,796 & 378 & 69.45 & 1.21692 & 1.0485 
    & 32.7 & 98.7 & 84.41 & 1.25851 & 1.3674 
    & 67.8 & 99.7 & 92.10 & \textbf{0.97371} & \textbf{1.0006}\\ \hline
\end{tabular}
}
\end{table*}

\subsection{Implementation Details}
We implement the proposed method under the ORB-SLAM2 framework~\cite{murartal17orbslam2}. 
In detail, our method is executed before every local BA in the local mapping part of the ORB-SLAM2. Instead of using the original implementation of the ORB-SLAM2, we modify it in two ways. Firstly, we change the existing multi-threaded processing to a single threaded processing for 1) an objective evaluation of the total runtime, 2) deterministic performance evaluation, and 3) disabling frame dropouts in optimization steps due to the processing delays.
Secondly, since the condition for deciding keyframes in ORB-SLAM2 depends on the number of tracking points from the local mapping thread and also on the state of the local mapping thread, the number of the local BA execution changes when the single threaded processing or point sparsification applied.
So, a deterministic keyframe insertion criterion that depends on the amount of change in translation and rotation is used to measure the impact of the proposed point sparsification.
Throughout the whole experiments, we set $\theta_{f}$ to the half of the edge's capacity. For solving the minimum-cost maximum-flow graph problem, we use the Google Optimization Research Tools~\cite{ortools}, which is an open-source, fast and portable software suite for solving combinatorial optimization problems.

\subsection{Evaluation Metric}
\label{section_metric}
We use the RMS of absolute trajectory error (ATE)~\cite{sturm12irosTUM} for pose accuracy measure, that implies the global consistency of the estimated trajectory by computing absolute distance between estimated pose and the ground truth pose. 
Since the concept of ATE, which represents the average deviation from the ground truth pose, is more suitable for evaluating the accuracy of the resulting visual map than relative pose error (RPE), which represents the local accuracy of trajectory, we evaluate rotational pose accuracy in the same sense as ATE and notate this as $ATE_r$. 
For the absolute trajectory error matrix at timestamp $i$, $E_i$, defined as
\begin{align}
E_i = {Q_i}^{-1}P_i,
\label{eq:ATE}
\end{align}
where $Q_i$ is the ground truth pose and $P_i$ is the estimated pose at timestamp $i$ aligned in the same coordinate frame,
RMS $ATE_r$ is computed as:
\begin{align}
ATE_r = \left({1 \over n} \sum_{i=1}^m {(|\angle rot(E_i)|)^2} \right)^{1 \over 2}
\label{eq:ATE_rotation}
\end{align}
based on the implementation in~\cite{grupp2017evo}.

\subsection{Performance Evaluation}
\label{section_total}
We evaluate the proposed method on the various datasets include EuRoC~\cite{Burri25012016}, TUM~\cite{sturm12irosTUM}, ScanNet~\cite{dai2017scannet} ICL-NUIM~\cite{handa14icl-nuim} and KITTI~\cite{geiger13KITTI}.
EuRoC is a popular indoor visual-inertial dataset collected from a Micro Aerial Vehicle (MAV) that contains synchronized stereo images, IMU measurements, and their ground truth poses. With 3-5 different trajectories for each three individual locations; Machine Hall (MH), Vicon Room1 (V1), and Vicon Room2 (V2), it offers a various sequences for both smooth or aggressive movements in small and large indoor environments. Here we use the EuRoC dataset for stereo implementation and do not use the IMU sensor data. Note that we exclude V2 data from our experiments because ORB-SLAM2 does not produce a good pose trajectory from the dataset due to its severe rotational movement, motion blurs, and lighting changes.
Table \ref{table_euroc} and Fig. \ref{figure_graph} shows the experimental results on EuRoC dataset in stereo mode.
The number of map points is soft-limited under the $M$ parameter. In the case of $M=100$, the reduction of map points reaches an average of 76\% for MH, which is a large place where about 20,000 map points are generated on average in original mode. For V1, which is a smaller place, there is a reduction of 60\% in the number of map points. 
At $M=200$ and $300$, a small amount of reduction is observed. The number of keyframes also decreases by up to 66\% because the reduction of map points that they have connections causes automatic dropouts of the keyframes whose connectivity falls under the threshold.

\begin{table*}[]
\centering
\caption{Comparison with ANMS point selection including the result of original. The lowest RMS ATE are in bold.}
\label{comparison_table}
\resizebox{\textwidth}{!}{%
\begin{tabular}{l|rrrr|rrrlr|rrrlr}
\hline
 & \multicolumn{4}{c|}{Original} & \multicolumn{5}{c|}{ANMS} & \multicolumn{5}{c}{Ours} \\ \cline{2-15} 
Dataset & \multicolumn{1}{l}{\#MPs} & \multicolumn{1}{l}{\#KFs} & \multicolumn{1}{l}{time (s)} & \multicolumn{1}{l|}{ATE(m)} & \multicolumn{1}{l}{MP (\%)} & \multicolumn{1}{l}{KF (\%)} & \multicolumn{1}{l}{time (s)} & connect. & \multicolumn{1}{l|}{ATE (m)} & \multicolumn{1}{l}{MP (\%)} & \multicolumn{1}{l}{KF (\%)} & \multicolumn{1}{l}{time (s)} & connect. & \multicolumn{1}{l}{ATE (m)} \\ \hline\hline
EuRoC & 16,815 & 293 & 197.06 & 0.06582 & 63.0 & 119.4 & 91.19 & 2.63 & 0.07709 & 61.7 & 79.8 & 144.46 & 6.72 & \textbf{0.05656} \\
TUM & 6,668 & 115 & 61.65 & 0.01831 & 41.7 & 102.8 & 45.79 & 4.90 & 0.03044 & 41.4 & 68.0 & 47.17 & 8.14 & \textbf{0.01779} \\
ICL-NUIM & 4,569 & 66 & 27.73 & 0.02016 & 72.0 & 103.7 & 24.23 & 6.54 & \textbf{0.01317} & 72.0 & 74.5 & 24.59 & 8.27 & 0.01565 \\
ScanNet & 24,945 & 507 & 499.11 & 0.06923 & 12.2 & 17.2 & 57.86 & 3.59 & 0.06970 & 12.3 & 14.8 & 100.57 & 6.96 & \textbf{0.04810} \\ \hline
\end{tabular}%
}
\end{table*}

Accordingly, the overall processing time is also greatly reduced since there is a significant gain in time consumption of bundle adjustment and tracking process by reducing the number of constraints. Fig. \ref{figure_graph} shows a graph of the total runtime and the RMS ATE for the experiments in table \ref{table_euroc}. 
The results show that our proposed method reduces the runtime to approximately a third while the performance is nearly maintained or even improved.
In the case where the error has increased compare to the original, the difference in ATE is within 4 cm at most.
For $M=300$, points/frames/time are mildly reduced by around 20\% , but it outperforms the original for all the sets. It can be interpreted as the proposed method successfully deletes only the points that negatively contributes to the optimization by unbalancing the constraints belonging to a pose.
Then, we evaluate the performance of the proposed method in various camera configurations using a subset of multifarious RGB-D indoor datasets, including TUM, ScanNet, and ICL-NUIM. TUM provides color and depth images, accelerometer data from Microsoft Kinect sensors, along with the sensor's true trajectories. We use 6 sequences from TUM, which are frequently used in other studies. 
ScanNet is a large-scale RGB-D dataset that provides detailed labels for a variety of tasks including 3D object classification, instance-level semantic segmentation. Its diverse, realistic environment and accurate ground truth makes it a high-quality benchmark for Visual SLAM. 
ICL-NUIM dataset contains color, depth images and ground measurements for two different synthetic scenes (living room and office scene). We use all sequences from ICL-NUIM. Table \ref{table_rgbd} shows the result for RGB-D datasets across multiple $M$ values. The range of $M$ is set differently because of the different spatial scale and fewer map points generated compared to the EuRoC dataset.
For $M=100$, the map points are reduced to an average of 40\%, and the number of keyframes is reduced to 60\% and the time to 74\%.
At the same time, the average ATE and $ATE_r$ decreases compared to the original, and even when it is higher than the original, the difference did not exceed 1cm and 0.6$^{\circ}$ at most. Also for $M=140$, $200$, the error decreases while taking the memory and time gain as well.

Fig. \ref{figure_door} is the visualization of saved map points on scene0000 of ScanNet together with 3d reconstructed environment of the scene obtained from estimated trajectory. In the original map, areas with disproportionately dense distribution of points are observed such as tile joints on floor, carpet patterns, and sofas. On the same environment, SLAM with proposed sparsification method makes visible differences. In the dense areas mentioned above, the number of point is drastically reduced with spatially even selection of points and therefore there are hardly few regions which have clusters of points. 
We observe a small number of points in some regions of the original map, such as the trash can, refrigerator, or the door because the texture is weak and the number of poses viewing those region is small. With the proposed method, those points are preserved well, so the region is not abandoned when tracking or localizing the scene.
Therefore, in the setting that provides robust performance for the original, it does not fail or is not significantly degraded even under a low-texture environment. But, it is worth considering a moderate sparsification by adjusting the capacity $M$ appropriately or adjusting the sparsification interval according to the characteristics of the visual scene.

\begin{figure}[th!]
\centering
\includegraphics[width=8cm]{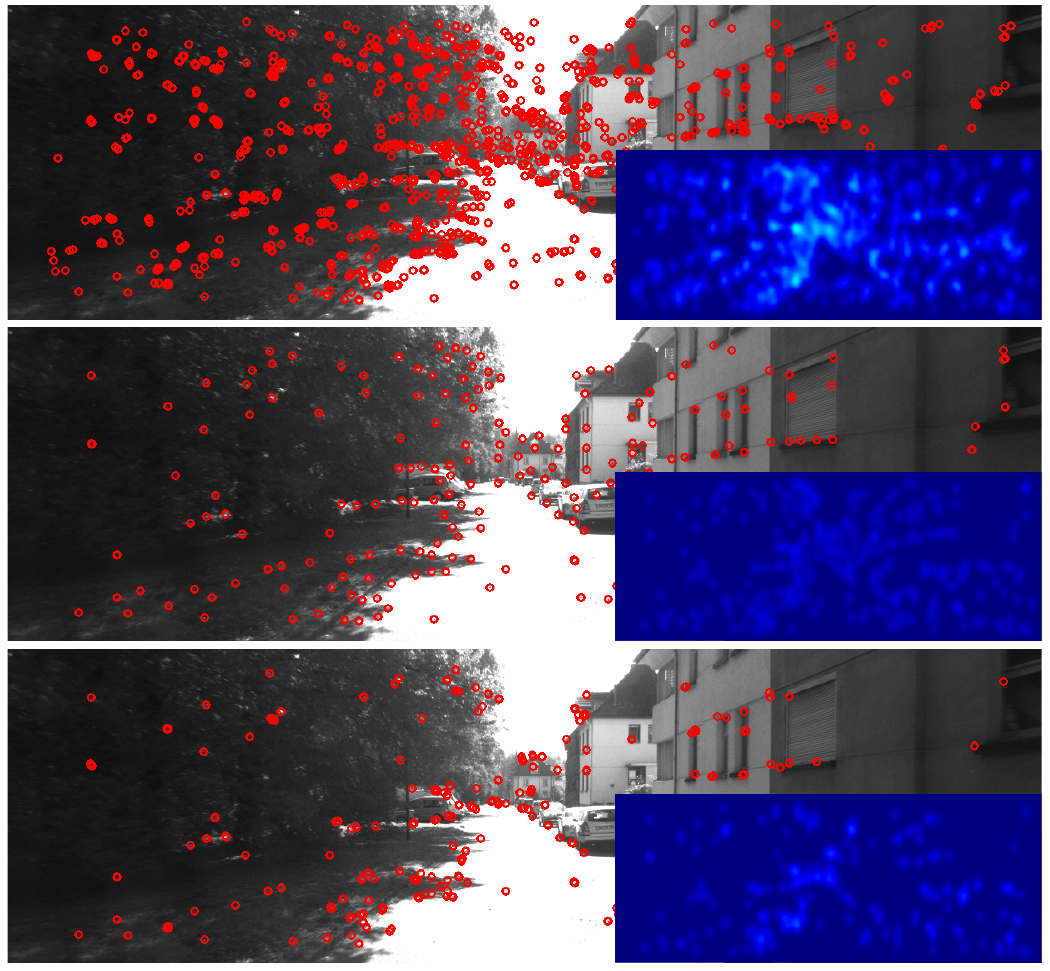}
  \caption{Visualization of selected keypoints on the image. From top to bottom, original, ANMS, ours. the lower right image shows the coverage and clusterity of keypoints.}
  \label{figure_image}
\end{figure}
We use the KITTI dataset for outdoor experiments. The table\ref{table_kitti} shows the experimental results with $M=100$, $200$ for three sequences 07, 08, and 10.
Unlike the indoor environment, which takes many map points and multiple connections in a relatively small scale of space and have many re-viewed points, keyframe dropping does not occur frequently in the outdoor environments. The time gain is also not as significant as on indoor. However, a performance improvement is also observed when a map point selection is performed conservatively like other results on the indoor.

\begin{table*}[t!]
\centering
\caption{Ablations of pose accuracy with a partial and the full costs on TUM dataset. The higher $C$, $F$, and $S$, the better. The lower ATE, the better. The best values across all the costs per each data sequence are in bold.}
\label{table_ablation}
\resizebox{\textwidth}{!}{%
\begin{tabular}{c|rrrr|rrrr|rrrr|rrrr}
\hline
\multicolumn{1}{l|}{} & \multicolumn{4}{c|}{All Costs} & \multicolumn{4}{c|}{$c_c$} & \multicolumn{4}{c|}{$c_c + c_b$} & \multicolumn{4}{c}{$c_c + c_s$} \\ \hline
Sequence & \multicolumn{1}{c}{C} & \multicolumn{1}{c}{F} & \multicolumn{1}{c}{S} & \multicolumn{1}{c|}{ATE (m)} & \multicolumn{1}{c}{C} & \multicolumn{1}{c}{F} & \multicolumn{1}{c}{S} & \multicolumn{1}{c|}{ATE (m)} & \multicolumn{1}{c}{C} & \multicolumn{1}{c}{F} & \multicolumn{1}{c}{S} & \multicolumn{1}{c|}{ATE (m)} & \multicolumn{1}{c}{C} & \multicolumn{1}{c}{F} & \multicolumn{1}{c}{S} & \multicolumn{1}{c}{ATE (m)} \\ \hline \hline
fr1\_desk & \multicolumn{1}{r}{7.50} & \multicolumn{1}{r}{709.67} & \multicolumn{1}{r}{2.89} & \textbf{0.01699} & \multicolumn{1}{r}{\textbf{7.74}} & \multicolumn{1}{r}{\textbf{757.76}} & \multicolumn{1}{r}{2.99} & 0.01824 & \multicolumn{1}{r}{6.46} & \multicolumn{1}{r}{743.50} & \multicolumn{1}{r}{2.97} & 0.01774 & \multicolumn{1}{r}{7.48} & \multicolumn{1}{r}{649.84} & \multicolumn{1}{r}{\textbf{3.14}} & 0.01790 \\ \hline
fr1\_desk2 & \multicolumn{1}{r}{5.20} & \multicolumn{1}{r}{738.93} & \multicolumn{1}{r}{\textbf{2.92}} & 0.02484 & \multicolumn{1}{r}{\textbf{5.23}} & \multicolumn{1}{r}{815.58} & \multicolumn{1}{r}{2.86} & 0.02651 & \multicolumn{1}{r}{4.78} & \multicolumn{1}{r}{\textbf{820.57}} & \multicolumn{1}{r}{2.85} & 0.02574 & \multicolumn{1}{r}{5.07} & \multicolumn{1}{r}{550.09} & \multicolumn{1}{r}{2.84} & \textbf{0.02476} \\ \hline
fr2\_desk & \multicolumn{1}{r}{4.47} & \multicolumn{1}{r}{586.20} & \multicolumn{1}{r}{2.95} & \textbf{0.01624} & \multicolumn{1}{r}{\textbf{4.55}} & \multicolumn{1}{r}{197.01} & \multicolumn{1}{r}{2.99} & 0.03665 & \multicolumn{1}{r}{3.51} & \multicolumn{1}{r}{\textbf{596.59}} & \multicolumn{1}{r}{2.78} & 0.01866 & \multicolumn{1}{r}{4.22} & \multicolumn{1}{r}{584.67} & \multicolumn{1}{r}{\textbf{3.00}} & 0.01842 \\ \hline
f2\_xyz & \multicolumn{1}{r}{20.62} & \multicolumn{1}{r}{2524.29} & \multicolumn{1}{r}{3.91} & 0.00785 & \multicolumn{1}{r}{\textbf{21.10}} & \multicolumn{1}{r}{2422.21} & \multicolumn{1}{r}{3.75} & 0.00787 & \multicolumn{1}{r}{19.73} & \multicolumn{1}{r}{\textbf{2560.01}} & \multicolumn{1}{r}{3.26} & \textbf{0.00743} & \multicolumn{1}{r}{19.52} & \multicolumn{1}{r}{2474.44} & \multicolumn{1}{r}{\textbf{3.99}} & 0.00843 \\ \hline
fr3\_office & \multicolumn{1}{r}{4.98} & \multicolumn{1}{r}{349.94} & \multicolumn{1}{r}{\textbf{3.42}} & \textbf{0.01347} & \multicolumn{1}{r}{4.98} & \multicolumn{1}{r}{374.26} & \multicolumn{1}{r}{3.40} & 0.02941 & \multicolumn{1}{r}{3.89} & \multicolumn{1}{r}{\textbf{359.33}} & \multicolumn{1}{r}{3.32} & 0.04014 & \multicolumn{1}{r}{\textbf{4.99}} & \multicolumn{1}{r}{354.03} & \multicolumn{1}{r}{3.39} & 0.01443 \\ \hline
fr3\_nt & \multicolumn{1}{r}{\textbf{7.97}} & \multicolumn{1}{r}{142.74} & \multicolumn{1}{r}{3.67} & \textbf{0.01533} & \multicolumn{1}{r}{7.83} & \multicolumn{1}{r}{144.20} & \multicolumn{1}{r}{\textbf{3.79}} & 0.02112 & \multicolumn{1}{r}{7.19} & \multicolumn{1}{r}{147.13} & \multicolumn{1}{r}{3.72} & 0.02225 & \multicolumn{1}{r}{\textbf{7.97}} & \multicolumn{1}{r}{\textbf{240.08}} & \multicolumn{1}{r}{3.77} & 0.01924 \\ \hline \hline
Avg. & \multicolumn{1}{r}{8.46} & \multicolumn{1}{r}{841.96} & \multicolumn{1}{r}{3.29} & \textbf{0.01579} & \multicolumn{1}{r}{\textbf{8.57}} & \multicolumn{1}{r}{785.17} & \multicolumn{1}{r}{3.30} & 0.02330 & \multicolumn{1}{r}{7.59} & \multicolumn{1}{r}{\textbf{871.19}} & \multicolumn{1}{r}{3.15} & 0.02199 & \multicolumn{1}{r}{8.21} & \multicolumn{1}{r}{808.86} & \multicolumn{1}{r}{\textbf{3.35}} & 0.01720 \\ \hline
\end{tabular}%
}
\end{table*}

\subsection{Comparisons}
\label{section_comparisons}
Adaptive non-maximal suppression (ANMS) algorithm~\cite{bailo18anms} improves the performance in SLAM and image registration by selecting keypoints detected on the image to be homogeneously distributed through efficient computation ~\cite{gauglitz11,brown2005multi}. 
ANMS achieves better results on visual SLAM compared to the topM~\cite{rosten2006machine} or the bucketing~\cite{kitt2010visual} approach when a sufficient number of keypoints is selected. 
We investigate the effect of selecting points with a high priority over spatial distribution when the number of points is significantly reduced as we propose here. Table~\ref{comparison_table} shows the result when ANMS is applied at the lowest ratio of point selection and when our method is applied according to the total number of map points produced by the ANMS.
The result is shown as the average value of each of the four datasets used in section \ref{section_total}, and the term {\textit{connect.}} refers to the average number of connections of each point. Our method shows better result for 3 out of 4 cases than ANMS as well as original.
Fig. \ref{figure_image} is a visualization of selected local interest points in one image by ANMS and our method. The points selected by ANMS are uniformly distributed as shown. However in the process of selecting a limited number of points, considering spatial distribution only derives to dropping points with high connectivity when given a point with a remote and low connectivity. Therefore, the overall point connectivity becomes weaker. This can also be confirmed with the number of keyframes, which our method has significantly fewer keyframes compared to ANMS while the number of map points is nearly identical since we keep the process of automatically culling keyframes with less than a certain amount of connections in ORB-SLAM2.
This means that only the map point with the main connection and the keyframes that observes it are preserved, and the keyframe that only sees the point that is not noticed by other frames is decimated by the proposed method.
As a result, it not only benefits significantly in the memory gain when considering memory consumption of keyframes is much higher since they contain a lot more information than map points, but also reduces the cost of bundle adjustment significantly by prunning numerous matches between points and frames. In the case of ICL-NUIM, the only dataset where ANMS perform better, the data is acquired in a small-scale space with rarely re-viewing the same area, moving cameras with almost only rotational movement while the position is fixed in the center of the space.
As a result, the difference in {\textit{connect.}} is not large compared to the other sets, and matches can be managed efficiently simply by evenly distributing points in the space.
Through this experiment, we claim that the connectivity is a key factor in the task of reducing the number of points efficiently to maintain the performance.

\subsection{Ablation Study}
We evaluate the effectiveness of the three costs proposed in Section \ref{section_total}:
$c_{c}$ is for maximizing the pose connectivity, $c_{s}$ is for maximizing the spatial diversity, and $c_{b}$ is for maximizing the pose baseline length of the keyframe pair. 
To see the effectiveness of each cost, we provide additional experiments conducted on TUM dataset in Table~\ref{table_ablation}. In particular, we consider the four scenarios: all costs, $c_{c}$ only, $c_{c}$ + $c_{s}$, and $c_{c}$ + $c_{b}$ because the major constraining cost is $c_{c}$ as mentioned in Section \ref{section_comparisons}. For each sequence, we observe the following attributes in addition to ATE:
\begin{itemize}
    \item $C$: Average number of connections with keyframes per point
    \item $F$: Maximum sequential difference between connected keyframes on point
    \item $S$: Average percentage of spatial occupancy by points on the image grid of the size $64$x$48$
\end{itemize}

As presented in Table~\ref{table_ablation}, when $c_{s}$ is combined with $c_{c}$, the ATE reduction is significantly larger than the case of $c_{c}$ only. In addition, compared with $c_{c}$, the performance of $c_{c}$ + $c_{s}$ is improved by a much larger difference than for $c_{c}$ + $c_{b}$.
With this, we can observe that the spatial distribution largely contribute on error reduction as other studies have shown.
$c_{c}$ + $c_{b}$ contributes to the selection of strong features and keeps the connection among multiple frames, so the spatial diversity decreases compared to the case of $c_{c}$ and the frame sequence interval is maximized.
When using all of these three costs, the lowest ATE is achieved by making more use of frames at a large baseline while maximizing the pose connectivity and the spatial diversity.

\begin{table}[h!]
\centering
%\caption{Localization Accuracy Evaluation on Sparsified Map}
\caption{Localization Accuracy Evaluation}
\label{table_localization}
\begin{tabular}{l|l|r|r}
\hline
\begin{tabular}[c]{@{}l@{}}Ref. Map\\  / Query Seq.\end{tabular} &  & \multicolumn{1}{l|}{Original} & \multicolumn{1}{c}{Ours ($M=100$)} \\ \hline\hline
\multirow{4}{*}{\begin{tabular}[c]{@{}l@{}}MH01\\ /MH02\end{tabular}} & time (s) & 97.08 & 66.41 (68.4\%) \\
 & \#MPs & 19,534 & 4,681 (23.9\%) \\
 & \#KFs & 398 & 137 (34.4\%) \\
 & ATE (m) & 0.03464 & \textbf{0.03105} \\ \hline
\multirow{4}{*}{\begin{tabular}[c]{@{}l@{}}scene0000\_00\\ /scene0000\_01\end{tabular}} & time (s) & 137.59 & \multicolumn{1}{l}{124.01 (91.1\%)} \\
 & \#MPs & 24,285 & 9,362 (38.6\%) \\
 & \#KFs & 464 & 274 (59.1\%) \\
 & ATE (m) & 0.07729 & \textbf{0.07643} \\ \hline
\end{tabular}
\end{table}  

\subsection{Localization Test}

The point sparsification capability and the pose trajectory estimation performance of the proposed method are shown in previous sections. 
Going further, we examine the pose localization accuracy against the original map and the sparsified map with two sets of sequences collected from the same scenes. 
We use MH01 \& MH02 from EuRoC dataset and scene0000\_00 \& scene0000\_01 from ScanNet. 
MH01 and scene0000\_00 are used to build the reference map in two modes, the original and the sparsified for conducting localization with MH02, scene0000\_01.
The localization results of the sequences MH02 and scene0000\_01 against the two maps are shown in the Table~\ref{table_localization}. In the case of MH01, the sparsified map only contains 23.9\% of points and 34.4\% of keyframes compared to the original map. The total localization time is reduced down to 68.4\% due to the gain in the map loading time, the initial position searching time, and the matching time. 
Despite such significant reductions both in the computation time and the map size, the RMS ATE calculated on every frame pose of query sequence MH02 even decreases. 
Similarly, scene0000 uses only 38.6\% of points and 59.1\% of keyframes compared to the original map, and the pose error decreases too.

\section{CONCLUSIONS}
We introduce a graph based point sparsification method for SLAM. The proposed method achieves three goals simultaneously during point sparsification: maximizing the point connectivity, maximizing the spatial diversity, and maximizing frame baseline length. With the baseline of ORB-SLAM2, our proposed method provides far more reduced map size while maintaining or even improving the pose tracking accuracy in the local mapping process. We conducted extensive evaluations and demonstrated that the proposed method can efficiently build a feature map in a significantly reduced size and other frames can be accurately localized against the sparsified map.
Our proposed method can be used for post-compression after map creation or for pre-processing before global bundle adjustment. It is generally applicable to local feature base SLAM systems, including multi-sensor SLAM, and provides an effective map decimation and a speedup to be applicable to other computationally challenging environments like wearable devices. 
Future directions of this research include a marginal graph optimization for fast optimization, and considering spatial density of the 3D points in addition to the 2D feature diversity.

\section*{ACKNOWLEDGMENT}
This work is supported in part by the Korea Agency for Infrastructure Technology Advancement (KAIA) grant funded by the Ministry of Land, Infrastructure and Transport (Grant 21AMDP-C160637-01).

%%%%%%%%% REFERENCES
{\small
\bibliographystyle{IEEEtranS}
\bibliography{references}

\begin{thebibliography}{10}
\providecommand{\url}[1]{#1}
\csname url@rmstyle\endcsname
\providecommand{\newblock}{\relax}
\providecommand{\bibinfo}[2]{#2}
\providecommand\BIBentrySTDinterwordspacing{\spaceskip=0pt\relax}
\providecommand\BIBentryALTinterwordstretchfactor{4}
\providecommand\BIBentryALTinterwordspacing{\spaceskip=\fontdimen2\font plus
\BIBentryALTinterwordstretchfactor\fontdimen3\font minus
  \fontdimen4\font\relax}
\providecommand\BIBforeignlanguage[2]{{%
\expandafter\ifx\csname l@#1\endcsname\relax
\typeout{** WARNING: IEEEtran.bst: No hyphenation pattern has been}%
\typeout{** loaded for the language `#1'. Using the pattern for}%
\typeout{** the default language instead.}%
\else
\language=\csname l@#1\endcsname
\fi
#2}}

\bibitem{ortools}
``Google operations research tools,'' \url{https://github.com/google/or-tools}.

\bibitem{bailo18anms}
O.~Bailo, F.~Rameau, K.~Joo, J.~Park, O.~Bogdan, and I.~S. Kweon, ``Efficient
  adaptive non-maximal suppression algorithms for homogeneous spatial keypoint
  distribution,'' \emph{Pattern Recognition Letters}, pp. 53--60, 2018.

\bibitem{brown2005multi}
M.~Brown, R.~Szeliski, and S.~Winder, ``Multi-image matching using multi-scale
  oriented patches,'' in \emph{2005 IEEE Computer Society Conference on
  Computer Vision and Pattern Recognition (CVPR'05)}, vol.~1.\hskip 1em plus
  0.5em minus 0.4em\relax IEEE, 2005, pp. 510--517.

\bibitem{Burri25012016}
M.~Burri, J.~Nikolic, P.~Gohl, T.~Schneider, J.~Rehder, S.~Omari, M.~W.
  Achtelik, and R.~Siegwart, ``The euroc micro aerial vehicle datasets,''
  \emph{Int. J. of Robot. Res.}, 2016.

\bibitem{cheng2006determining}
Z.~Cheng, D.~Devarajan, and R.~J. Radke, ``Determining vision graphs for
  distributed camera networks using feature digests,'' \emph{EURASIP Journal on
  Advances in Signal Processing}, vol. 2007, pp. 1--11, 2006.

\bibitem{choudhary15}
S.~Choudhary, V.~Indelman, H.~I. Christensen, and F.~Dellaert,
  ``Information-based reduced landmark slam,'' in \emph{Int. Conf. Robots and
  Automation (ICRA)}, 2015.

\bibitem{concha19}
E.~K.~T. Concha, D.~Pittol, R.~Westhauser, M.~Kolberg, R.~Maffei, and
  E.~Prestes, ``Map point optimization in keyframe-based {SLAM} using
  covisibility graph and information fusion,'' in \emph{Int. Conf. Advanced
  Robotics (ICAR)}, 2019.

\bibitem{cvisic15}
I.~Cvišić and I.~Petrović, ``Stereo odometry based on careful feature
  selection and tracking,'' in \emph{2015 European Conference on Mobile Robots
  (ECMR)}, 2015, pp. 1--6.

\bibitem{dai2017scannet}
A.~Dai, A.~X. Chang, M.~Savva, M.~Halber, T.~Funkhouser, and M.~Nie{\ss}ner,
  ``Scannet: Richly-annotated 3d reconstructions of indoor scenes,'' in
  \emph{CVPR}, 2017.

\bibitem{dias19}
N.~Dias and G.~Laureano, ``Accurate stereo visual odometry based on keypoint
  selection,'' in \emph{2019 Latin American Robotics Symposium (LARS), 2019
  Brazilian Symposium on Robotics (SBR) and 2019 Workshop on Robotics in
  Education (WRE)}, 2019, pp. 74--79.

\bibitem{edmonds72}
J.~Edmonds and R.~Karp, ``Theoretical improvements in algorithmic efficiency
  for network flow problems,'' \emph{Journal of the Association for Computing
  Machinery}, vol.~19, no.~2, p. 248–264, 1972.

\bibitem{fanfani16}
M.~Fanfani, F.~Bellavia, and C.~Colombo, ``Accurate keyframe selection and
  keypoint tracking for robust visual odometry,'' \emph{Machine Vision and
  Applications}, 08 2016.

\bibitem{fontan20}
A.~Fontán, J.~Civera, and R.~Triebel, ``Information-driven direct rgb-d
  odometry,'' in \emph{CVPR}, 2020.

\bibitem{gauglitz11}
S.~Gauglitz, L.~Foschini, M.~Turk, and T.~H{\"{o}}llerer, ``Efficiently
  selecting spatially distributed keypoints for visual tracking,'' in
  \emph{ICIP}, 2011.

\bibitem{geiger13KITTI}
A.~Geiger, P.~Lenz, C.~Stiller, and R.~Urtasun, ``Vision meets robotics: The
  {KITTI} dataset,'' \emph{Int. J. of Robot. Res.}, vol.~32, no.~11, pp.
  1231--1237, 2013.

\bibitem{goldberg1997efficient}
A.~V. Goldberg, ``An efficient implementation of a scaling minimum-cost flow
  algorithm,'' \emph{Journal of algorithms}, vol.~22, no.~1, pp. 1--29, 1997.

\bibitem{gruber2010optimization}
L.~Gruber, S.~Zollmann, D.~Wagner, D.~Schmalstieg, and T.~Hollerer,
  ``Optimization of target objects for natural feature tracking,'' in
  \emph{2010 20th International Conference on Pattern Recognition}.\hskip 1em
  plus 0.5em minus 0.4em\relax IEEE, 2010, pp. 3607--3610.

\bibitem{grupp2017evo}
M.~Grupp, ``evo: Python package for the evaluation of odometry and slam.''
  \url{https://github.com/MichaelGrupp/evo}, 2017.

\bibitem{handa14icl-nuim}
A.~Handa, T.~Whelan, J.~McDonald, and A.~Davison, ``A benchmark for {RGB-D}
  visual odometry, {3D} reconstruction and {SLAM},'' in \emph{Int. Conf. Robots
  and Automation (ICRA)}, 2014.

\bibitem{Lepetit17}
M.~H{\"o}ll and V.~Lepetit, ``Monocular lsd-slam integration within ar
  system,'' \emph{ArXiv}, vol. abs/1702.02514, 2017.

\bibitem{hsiung18}
J.~Hsiung, M.~Hsiao, E.~Westman, R.~Valencia, , and M.~Kaess, ``Information
  sparsification in visual-inertial odometry,'' in \emph{IEEE/RSJ Intl. Conf.
  on Intell. Robots and Syst. (IROS)}, 2018.

\bibitem{huang13}
G.~Huang, M.~Kaess, and J.~J. Leonard, ``Consistent sparsification for graph
  optimization,'' in \emph{European Conference on Mobile Robots}, 2013.

\bibitem{engel14}
T.~S. J.~Engel and D.~Cremers, ``Lsd-slam: Largescale direct monocular slam,''
  in \emph{ECCV}, 2014.

\bibitem{kitt2010visual}
B.~Kitt, A.~Geiger, and H.~Lategahn, ``Visual odometry based on stereo image
  sequences with ransac-based outlier rejection scheme,'' in \emph{2010 ieee
  intelligent vehicles symposium}.\hskip 1em plus 0.5em minus 0.4em\relax IEEE,
  2010, pp. 486--492.

\bibitem{klein07PTAM}
G.~Klein and D.~Murray, ``Parallel tracking and mapping for small ar
  workspaces,'' in \emph{2007 6th IEEE and ACM International Symposium on Mixed
  and Augmented Reality}, 2007, pp. 225--234.

\bibitem{frey18}
J.~P.~H. Kristoffer M.~Frey, Ted J.~Steiner, ``Complexity analysis and
  efficient measurement selection primitives for high-rate graph slam,'' in
  \emph{Int. Conf. Robots and Automation (ICRA)}, 2018.

\bibitem{lin19}
X.~Lin, F.~Wang, L.~Guo, and W.~Zhang, ``An automatic key-frame selection
  method for monocular visual odometry of ground vehicle,'' \emph{IEEE Access},
  vol.~7, pp. 70\,742--70\,754, 2019.

\bibitem{murartal17orbslam2}
R.~Mur-Artal and J.~D. Tardós, ``{ORB-SLAM2}: an open-source {SLAM} system for
  monocular, stereo and {RGB-D} cameras,'' \emph{IEEE Trans. Robotics},
  vol.~33, no.~5, pp. 1255--1262, 2017.

\bibitem{paull19}
L.~Paull, G.~Huang, and J.~J. Leonard, ``A unified resource-constrained
  framework for graph {SLAM},'' in \emph{Int. Conf. Robots and Automation
  (ICRA)}, 2016.

\bibitem{rosten2006machine}
E.~Rosten and T.~Drummond, ``Machine learning for high-speed corner
  detection,'' in \emph{European conference on computer vision}.\hskip 1em plus
  0.5em minus 0.4em\relax Springer, 2006, pp. 430--443.

\bibitem{Schueftan15}
D.~S. Schueftan, M.~J. Colorado, and I.~F. Mondragon~Bernal, ``Indoor mapping
  using slam for applications in flexible manufacturing systems,'' in
  \emph{2015 IEEE 2nd Colombian Conference on Automatic Control (CCAC)}, 2015,
  pp. 1--6.

\bibitem{sturm12irosTUM}
J.~Sturm, N.~Engelhard, F.~Endres, W.~Burgard, and D.~Cremers, ``A benchmark
  for the evaluation of rgb-d slam systems,'' in \emph{IEEE/RSJ Intl. Conf. on
  Intell. Robots and Syst. (IROS)}, Oct. 2012.

\bibitem{qin20}
Y.~C. Tong~Qin, Tongqing~Chen and Q.~Su, ``Avp-slam: Semantic visual mapping
  and localization for autonomous vehicles in the parking lot,'' in
  \emph{IEEE/RSJ Intl. Conf. on Intell. Robots and Syst. (IROS)}, 2020.

\bibitem{vallve2019pose}
J.~Vallv{\'e}, J.~Sol{\`a}, and J.~Andrade-Cetto, ``Pose-graph slam
  sparsification using factor descent,'' \emph{Robotics and Autonomous
  Systems}, vol. 119, pp. 108--118, 2019.

\bibitem{vallve18}
J.~Vallvé, J.~Solà, and J.~Andrade-Cetto, ``Graph {SLAM} sparsification with
  populated topologies using factor descent optimization,'' 2018.

\bibitem{Opdenbosch18}
D.~Van~Opdenbosch, M.~Oelsch, A.~Garcea, T.~Aykut, and E.~Steinbach,
  ``Selection and compression of local binary features for remote visual
  slam,'' in \emph{2018 IEEE International Conference on Robotics and
  Automation (ICRA)}, 2018, pp. 7270--7277.

\bibitem{Stumberg17}
L.~von Stumberg, V.~Usenko, J.~Engel, J.~Stückler, and D.~Cremers, ``From
  monocular slam to autonomous drone exploration,'' in \emph{2017 European
  Conference on Mobile Robots (ECMR)}, 2017, pp. 1--8.

\bibitem{wang18AprilSAM}
X.~Wang, R.~Marcotte, G.~Ferrer, and E.~Olson, ``April{SAM}: Real-time
  smoothing and mapping,'' in \emph{Int. Conf. Robots and Automation (ICRA)},
  2018.

\bibitem{zhou20sba}
L.~Zhou, Z.~Luo, M.~Zhen, T.~Shen, S.~Li, Z.~Huang, T.~Fang, , and L.~Quan,
  ``Stochastic bundle adjustment for efficient and scalable 3d
  reconstruction,'' in \emph{Eur. Conf. Comput. Vis. (ECCV)}, 2020.

\end{thebibliography}
}
\end{document}